\documentclass{article}

\PassOptionsToPackage{numbers,sort}{natbib}

\usepackage[preprint]{neurips_2019}

\usepackage[utf8]{inputenc} 
\usepackage[T1]{fontenc}    
\usepackage{hyperref}       
\usepackage{url}            
\usepackage{booktabs}       
\usepackage{amsfonts}       
\usepackage{nicefrac}       
\usepackage{microtype}      
\usepackage{graphicx}

\usepackage{amsmath}
\usepackage{amssymb}
\usepackage{subcaption}
\usepackage{adjustbox}
\usepackage{threeparttable}
\usepackage{multirow}
\usepackage{algorithm}
\usepackage{algorithmicx}
\usepackage{algpseudocode}
\usepackage{enumitem}

\usepackage{color}
\definecolor{darkred}{rgb}{0.8,0,0}
\definecolor{darkgreen}{rgb}{0,0.7,0}
\definecolor{darkblue}{rgb}{0,0.0,0.3}
\hypersetup{
  colorlinks,
  linkcolor=darkred,
  citecolor=darkgreen,
  urlcolor=darkblue,
  pdfinfo={
    Title={Federated Learning with Additional Mechanisms on Clients to Reduce Communication Costs},
    Author={Xin Yao, Tianchi Huang, Chenglei Wu, Rui-Xiao Zhang, Lifeng Sun},
  }
}

\title{Federated Learning with Additional Mechanisms on Clients to Reduce Communication Costs}

\author{
  Xin Yao, Tianchi Huang, Chenglei Wu, Rui-Xiao Zhang, Lifeng Sun\\
  Department of Computer Science and Technology \\
  Tsinghua University, Beijing, China \\
  \texttt{\{yaox16,htc19,wucl18,zhangrx17\}@mails.tsinghua.edu.cn} \\
  \texttt{sunlf@tsinghua.edu.cn}\\
}

\begin{document}

\maketitle

\begin{abstract}
Federated learning (FL) enables on-device training over distributed networks consisting of a massive amount of modern smart devices, such as smartphones and IoT~(Internet of Things) devices.
However, the leading optimization algorithm in such settings, i.e., \emph{federated averaging} (FedAvg), suffers from heavy communication costs and the inevitable performance drop, especially when the local data is distributed in a non-IID way.
To alleviate this problem, we propose two potential solutions by introducing additional mechanisms to the on-device training.

The first (FedMMD) is adopting a two-stream model with the MMD (Maximum Mean Discrepancy) constraint instead of a single model in vanilla FedAvg to be trained on devices.
Experiments show that the proposed method outperforms baselines, especially in non-IID FL settings, with a reduction of more than 20\% in required communication rounds.

The second is FL with feature fusion (FedFusion).
By aggregating the features from both the local and global models, we achieve higher accuracy at fewer communication costs.
Furthermore, the feature fusion modules offer better initialization for newly incoming clients and thus speed up the process of convergence.
Experiments in popular FL scenarios show that our FedFusion outperforms baselines in both accuracy and generalization ability while reducing the number of required communication rounds by more than 60\%.

\end{abstract}

\section{Introduction}
\label{sec:intro}

Mobile phones, wearable devices, and IoT~(Internet of Things) devices play an important role in modern life.
Intelligent applications on these devices are becoming popular, such as intelligent personal assistant, machine translation, keyboard input suggestion, etc.
These applications usually use pre-trained models and perform forward inference on clients, which lacks flexibility and personalization.
Meanwhile, smart edge devices are generating a tremendous amount of valuable yet privacy-sensitive data that has the potential to improve existing models.
To take full advantage of the on-device data, traditional machine learning strategies require collecting data from clients, training a centralized model on the servers and then issuing the model to clients, which puts a heavy burden on the communication networks and is exposed to high privacy risks.

Recently, a series of work called \emph{federated learning} (FL) \cite{konevcny2015federated,konevcny2016federated,mcmahan2017communication} enables on-device training directly over the distributed networks.
The aim of FL is to train a model from data $\{X^1,...,X^K\}$ generated by $K$ distributed clients (each device is treated as a client).
Each client, $t\in[K]$, generates data in a non-IID manner, which means the data distribution on client $t$,  $X^t \sim P^t$, is not a uniform sample of the whole distribution.
The leading algorithm in FL, i.e., \emph{federated averaging} (FedAvg) \cite{mcmahan2017communication}, assumes a synchronous updating scheme that proceeds in rounds of communication.
Considering a fixed number of $K$ clients (each with a local dataset), a random fraction $C$ of clients is selected to participate in this round of updating at the beginning of each round.
The server sends the current \emph{global model} to each of these chosen clients.
Then each client computes a unique \emph{local model} based on the global model and its local data, and reports it to the server.
Finally, the server updates the global model by model averaging and begins the next round.
By adding more computational iterations to clients, FedAvg reduces the required communication rounds compared with traditional SGD based methods.
However, further studies \cite{jeong2018communication,yao2018twostream} point out that communication costs remain the main constraint in FL compared to other factors, e.g., the computation costs, and the accuracy of FedAvg would drop significantly if the models were trained with pathological non-IID data.

To alleviate this problem, we propose two potential solutions by introducing additional mechanisms to the on-device training.

We first propose adopting a two-stream model, which is commonly used in transfer learning and domain adaptation \cite{sun2016deep,long2015learning,zhuo2017deep}, instead of a single model to be trained on devices in FL settings.
Maximum Mean Discrepancy (MMD) constraint \cite{gretton2012optimal} is introduce to the on-device training iterations of our method, which forces the local model to integrating more knowledge from the global one.
Further experiments show the proposed FedMMD brings a reduction in required communication rounds without affecting the final test performance.

Then we further propose a new FL algorithm with feature fusion mechanism (FedFusion) to reduce the communication costs.
By introducing the feature fusion modules, we aggregate the features from both the local and global models after the feature extraction stage with little extra computation costs.
These modules make the training process on each client more efficient and handle the non-IID data distribution more pertinently, as each client will learn the most appropriate feature fusion module for itself.

In conclusion, our contributions are as follows:
\begin{itemize}[leftmargin=*]
\item We use a two-stream model with MMD (Maximum Mean Discrepancy) constraint instead of a single model in vanilla FedAvg to be trained on devices.
\item We propose aggregating the features from both the local and global models during the on-device training.
\item Experiments on popular FL settings show that the proposed methods outperform baselines in both accuracy and generalization ability while reducing the number of communication rounds by up to more than 60\%.
\end{itemize}

\section{Related Work}
\label{sec:related}

\subsection{Federated Learning}

Federated learning~(FL) is proposed by McMahan et al. \cite{mcmahan2017communication} to tackle the problem of decentralized training over massively distributed intelligent devices without access to the privacy-sensitive data directly.

Considering that communication costs remain the main constraint in FL, some research efforts have already been made.
Kone{\v{c}}n{\`y} et al. \cite{konevcny2016federated} proposed structured and sketched updates in the context of client-to-server communication.
Yao et al. \cite{yao2018twostream} introduced extra constraints to the on-device training procedure, aiming to integrate more knowledge from other clients while fitting the local data.
Caldas et al. \cite{caldas2018expanding} proposed federated dropout to train subsets on clients and extended the lossy compression \cite{suresh2017distributed} to server-to-client communication.

\subsection{Maximum Mean Discrepancy}
\label{sec:mmd}

As the name suggests, Maximum Mean Discrepancy (MMD) measures the distance between the means of two data distributions, and is widely used in domain adaptation problems \cite{long2015learning,long2013transfer,tzeng2014deep}, describing the difference between features generated from source and target domains.
By minimizing the MMD loss, they force the two-stream model extract more generalized features, which is very similar to our purpose of learning better representations in the whole dataset. 
In this paper, we focus on the multiple kernel variant of MMD (MK-MMD) proposed by Gretton et al \cite{gretton2012optimal}, which first maps the data distributions to a Reproducing Kernel Hilbert Space (RKHS).

Given two data distributions $x$ and $y$, the square of MMD between them can be expressed as:
\begin{equation}
\mathit{MMD}^2(x, y) = \| E[\phi(x)] - E[\phi(y)] \|^2
\end{equation}
where $\phi(\cdot)$ denotes the mapping to RKHS.
In practice, this mapping is unknown.
Using the kernel trick to replace the inner product and we have:
\begin{equation}
\label{eq:mmd}
\mathit{MMD}^2(x, y) = E[K(x, x)] + E[K(y, y)] - 2E[K(x, y)]
\end{equation}
where $K(x, y) = \langle\phi(x), \phi(y)\rangle$ is the desired kernel function.
In this paper, we use a standard radial basis function (RBF) kernel with multiple width.

\section{Methods}
\label{sec:methods}

In this section, we will first introduce our two-stream FL with MMD (FedMMD), then introduce the proposed feature fusion modules and give our FL algorithm with feature fusion mechanism (FedFusion).

\begin{figure}[t]
    \includegraphics[width=\linewidth]{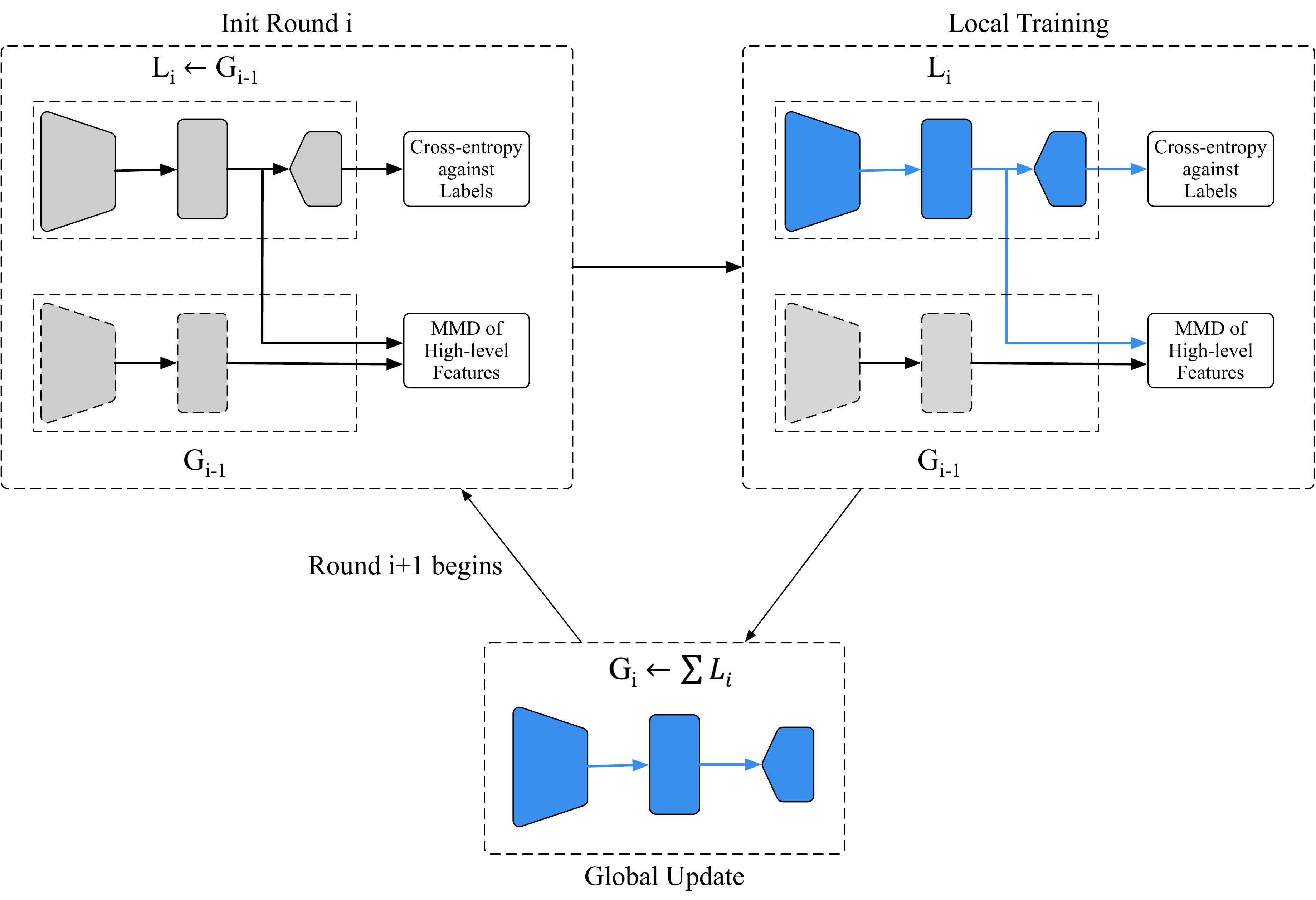}
	\caption{Two-stream model with MMD: the global model is fixed while the local model is trained through back propagation. (Better viewed in color)}
	\label{fig:fedmmd}
\end{figure}

\subsection{Two Stream Federated Learning}

Formally, let $\Theta^G$ and $\Theta^L$ be the model parameters, that is, the weights and biases, of all the layers in the global and local models respectively. 
Let $X^t = \{ x^t_i \}^{n_t}_{i=1}$ and $Y^t = \{ y^t_i \}^{n_t}_{i=1}$ denote the training data and the corresponding labels on client $t$, where $n_t$ is the number of examples.

Figure \ref{fig:fedmmd} shows how our FedMMD works on the specific client $t$.
As shown in the figure, the global model is received from the server at the beginning of the current round and fixed in the following training process, while the local model is initialize with the parameters of the global model ($\Theta^L \leftarrow \Theta^G$) and then trained on the local data $X^t$ and labels $Y^t$ by minimizing a loss function in the form:
\begin{align}
L(\Theta^L|\Theta^G, X^t, Y^t) &= L_{cls} + L_{\mathit{MMD}} \\
L_{cls} &= \frac{1}{n^t}\sum^{n^t}_{i=1}J(\theta^L(x^t_i), y^t_i) \\
L_{\mathit{MMD}} &= \lambda \mathit{MMD}^2(\theta^G(X^t), \theta^L(X^t))
\end{align}
where $\theta^G(X^t)$ and $\theta^L(X^t)$ denote the outputs of the global model $\Theta^G$ and local model $\Theta^L$ with the corresponding input $X^t$ respectively.
$J(\theta(x), y)$ denotes a standard classification loss, such as cross-entropy loss function in our experiments. $L_{\mathit{MMD}}$ denotes the MMD loss between the outputs of the global and local models computed by Equation (\ref{eq:mmd}).
This term is weighted by coefficient $\lambda$.

\begin{algorithm}[t]
  \caption{FedMMD}
  \label{algorithm1}
  \begin{algorithmic}[1]
    \Require
    \State initialize $\Theta^G_0$
    \For {each round r = 1, 2, ...}
    	\State $S_r \leftarrow$ (random set of $C \cdot K$ clients)
        \For {each client $t \in S_r$}
        	\State $\Theta^t_{r+1} \leftarrow Client(\Theta^G_r)$
        \EndFor
        \State $\Theta^G_{r+1} \leftarrow \sum_{t \in S_r} \Theta^t_{r+1}$
    \EndFor
  \end{algorithmic}
  \begin{algorithmic}[1]
    \Ensure Run on client $t$
    \For {each local epoch}
    	\For {each batch b}
        	\State Update $\Theta^L$ by minimizing $L(\Theta^L|\Theta^G, X^t, Y^t)$
        \EndFor
    \EndFor
    \State return $\Theta^L$ to server
  \end{algorithmic}
\end{algorithm}

The training process of FedAvg is indeed a cycle process of learning local representations, merging knowledge from different clients and then learning again.
In other words, the global model contains more knowledge from multiple clients while the local model learns better representations of the local data. 
Compared with a single local model trained in FedAvg, we keep the global model as a reference instead of throwing it away after initializing.
As discussed in Section \ref{sec:mmd}, MMD is a measurement of the distance between the means of two data distributions.
By minimizing the MMD loss term between the outputs of the global and local models, we force the local model to integrate more knowledge from other clients in addition to the representations of data on the current client, thus accelerating the convergence of the overall training process, in other words, reducing the communication rounds.

Our two-stream FL algorithm with MMD is described as Algorithm \ref{algorithm1}.

\subsection{Feature Fusion Modules}
\label{sec:fusion}

\begin{figure}[t]
\begin{subfigure}[b]{.32\linewidth}
\centering
\includegraphics[width=0.9\linewidth]{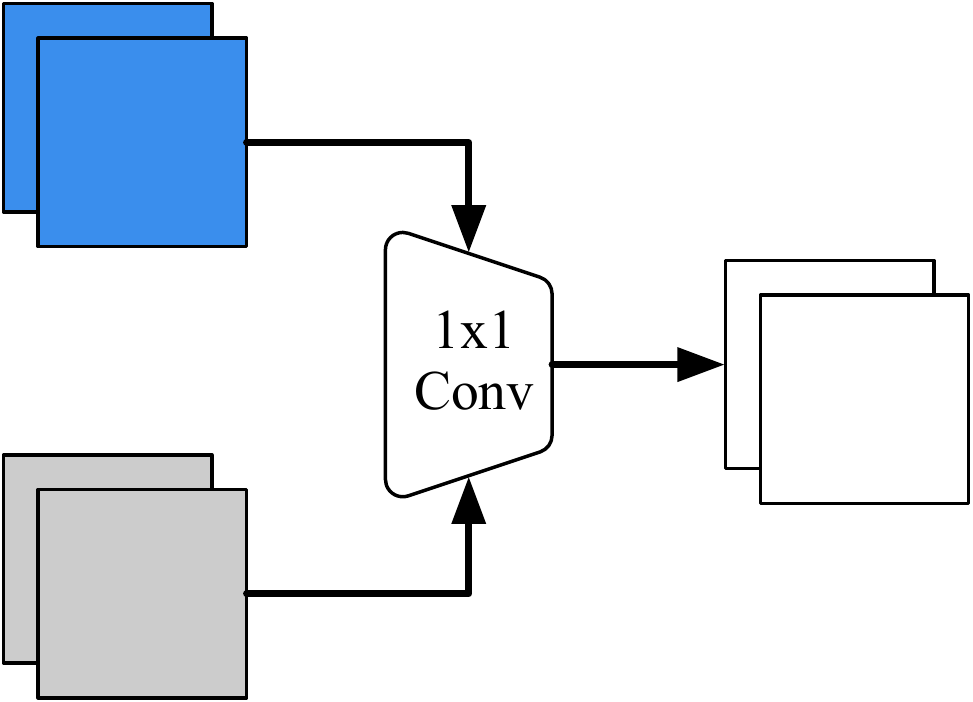}
\caption{\label{fig:a_conv} Conv}
\end{subfigure}
\begin{subfigure}[b]{.32\linewidth}
\centering
\includegraphics[width=0.9\linewidth]{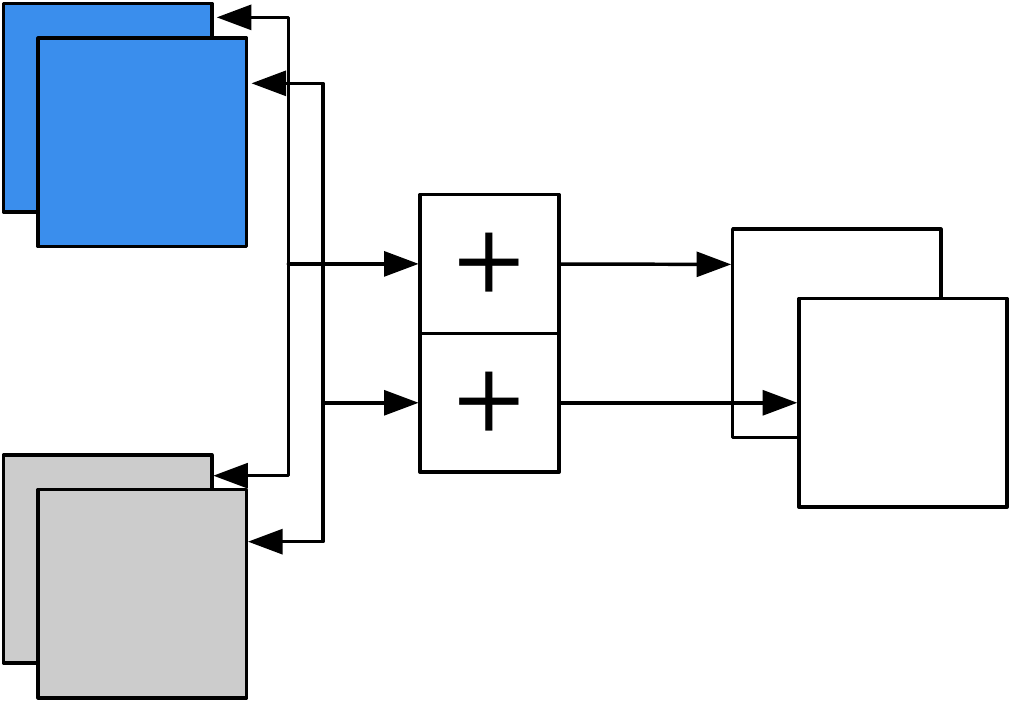}
\caption{\label{fig:b_multi} Multi}
\end{subfigure}
\begin{subfigure}[b]{.32\linewidth}
\centering
\includegraphics[width=0.9\linewidth]{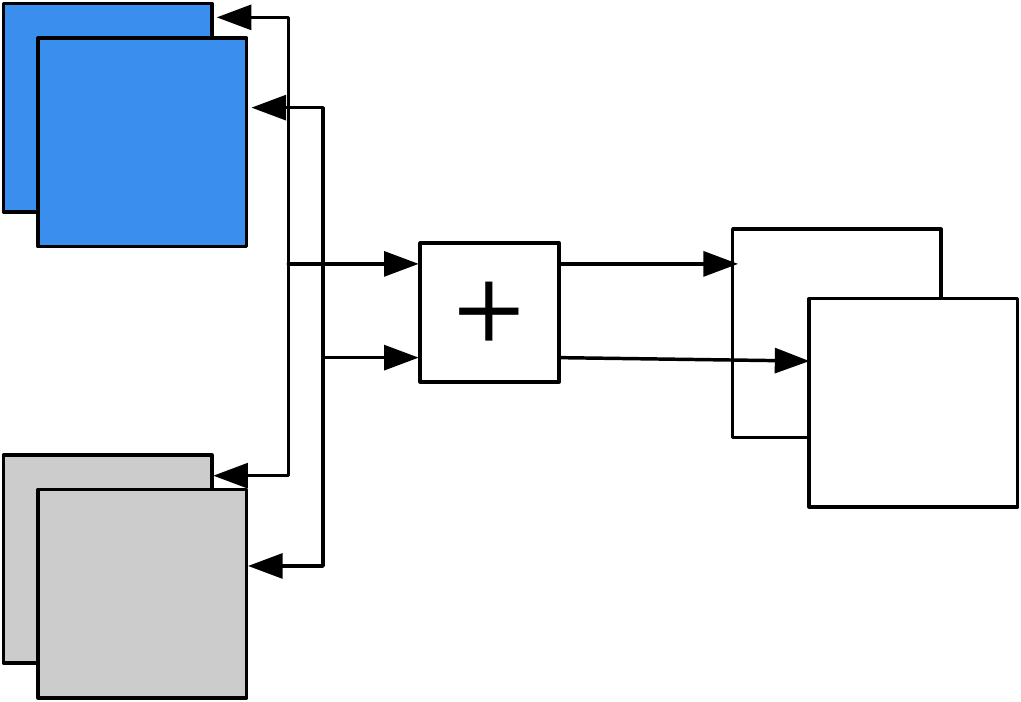}
\caption{\label{fig:c_single} Single}
\end{subfigure}
\caption{Three types of feature fusion modules. The fusion operator is actually a mapping function, $F: \mathbb{R}^{2C \times H \times W} \rightarrow \mathbb{R}^{C \times H \times W}$}
\label{fig:fusion}
\end{figure}

The detailed architectures of feature fusion modules are illustrated in Figure \ref{fig:fusion}.

Concretely, an input image $\pmb{x}$ is transformed into two feature spaces by the local feature extractor $E_l$ and the global one $E_g$ respectively, with the feature maps $E_l(x), E_g(x) \in \mathbb{R}^{C \times H \times W}$.
Then a fusion operator $F$ embeds $E_l(x)$ and $E_g(x)$ into a fusion feature space, where $F(E_l(x), E_g(x)) \in \mathbb{R}^{C \times H \times W}$.
In this paper, we introduce three types of fusion operator as follows.

\textit{Conv} operator ($F_{conv}$) is implemented with $1\times1$ convolutions,
\begin{align}
\label{eq:conv}
    F_{conv}(E_l(x), E_g(x)) = \pmb{\mathit{W}}_{conv}(E_g(x) || E_l(x))
\end{align}
where $\pmb{\mathit{W}}_{conv} \in \mathbb{R}^{2C \times C}$ is the learned weight matrix and $||$ denotes the operation that concatenates the feature maps along channel axis.

\textit{Multi} operator ($F_{multi}$) introduces a learned weight \textit{vector} $\pmb{\lambda} \in \mathbb{R}^C$ and computes the weighted sum between the local and global feature maps,
\begin{align}
\label{eq:multi}
    F_{multi}(E_l(x), E_g(x)) = \pmb{\lambda} E_g(x) + (\pmb{1} - \pmb{\lambda}) E_l(x)
\end{align}
where the weighted vector $\pmb{\lambda}$ is first broadcasted to the shape of $C \times H \times W$ and then multiplied by the feature maps element-wise, as illustrated in Figure \ref{fig:b_multi}.

\textit{Single} operator ($F_{single}$) uses a learned weight \textit{scalar} $\lambda$ and computes the weighted sum between the local and global feature maps,
\begin{align}
\label{eq:single}
    F_{single}(E_l(x), E_g(x)) = \lambda E_g(x) + (1 - \lambda) E_l(x)
\end{align}
where the global and local feature maps are scaled by $\lambda$ and $1-\lambda$ respectively and then added together element-wise, as shown in Figure \ref{fig:c_single}.

\begin{figure}
    \centering
    \includegraphics[width=\linewidth]{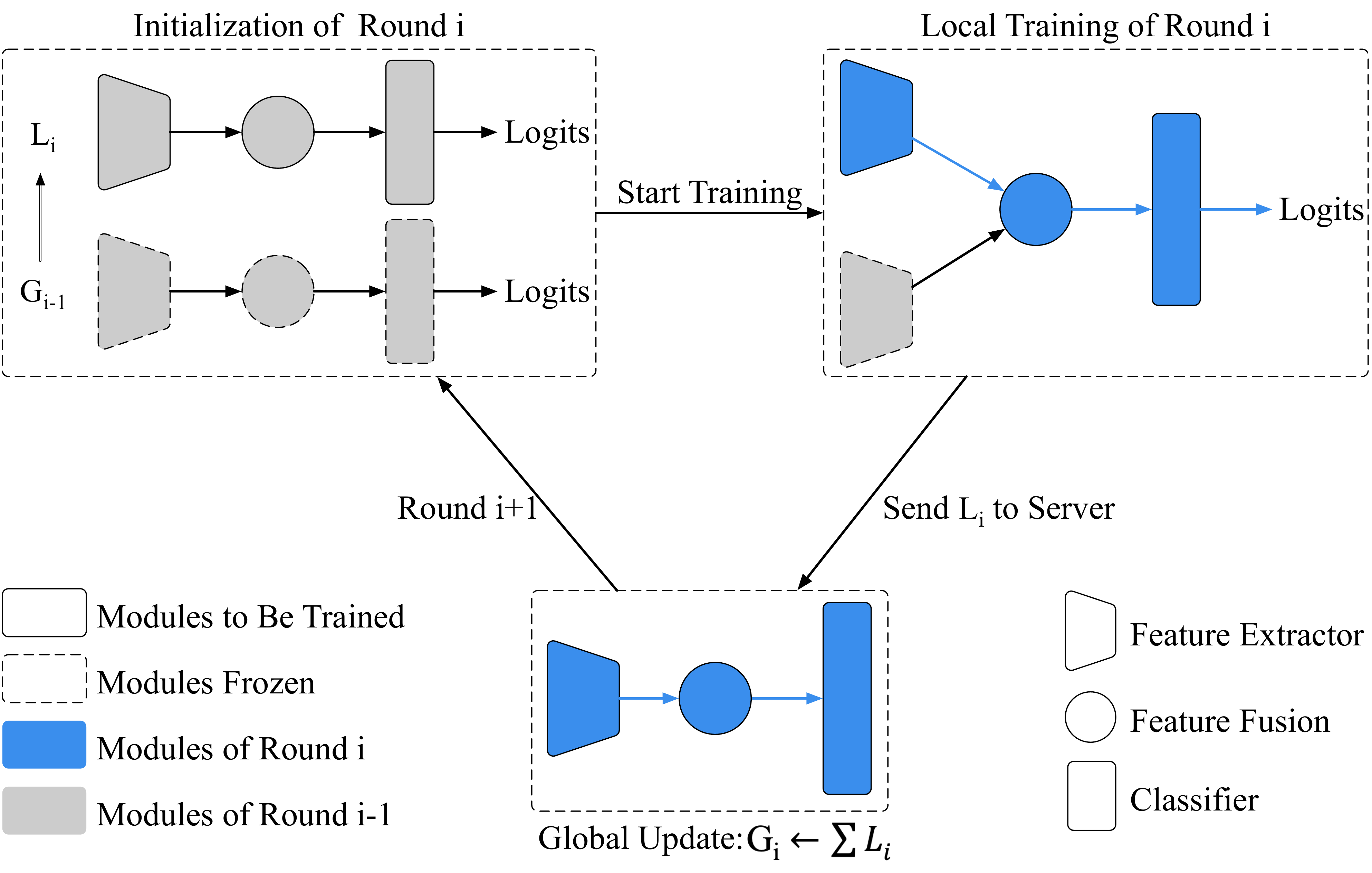}
    \caption{Training Iteration of FedFusion. During the training procedure on clients, the local and global feature extractors are concatenated by a \textit{feature fusion module}.}
    \label{fig:fedfusion}
\end{figure}

\begin{algorithm}[t]
\caption{FedFusion}
\label{algorithm}
\begin{algorithmic}[1]
    \Require
    \State initialize $G_0$
    \For {each round r = 0, 1, 2, ...}
        \State $S_r \leftarrow$ random sample $m$ clients
        \For {each client $t \in S_r$} in parallel
            \State $L^t_{r+1} \leftarrow \textbf{Client}(G_r)$
        \EndFor
        \State $G_{r+1} \leftarrow \frac{1}{n_{S_r}}\sum_{t \in S_r} n_t L^t_{r+1}$
    \EndFor
\end{algorithmic}
\begin{algorithmic}[1]
    \Ensure Round $r$ on client $t$
    \State $L^t_r = C \circ F \circ E_l \leftarrow G_r$ \indent// $C$ is the classifier
    \For {each batch $(x, y)$}
        \State Compute $\mathcal{L}(C \circ F(E_l(x), E_g(x)), y)$
        \State Update $E_l, F, C$ by backpropagation
    \EndFor
    \State return $L^t_{r+1}$ to server
\end{algorithmic}
\end{algorithm}

\subsection{Federated Learning with Feature Fusion Mechanism}

A typical training iteration of the proposed FedFusion is shown in Figure \ref{fig:fedfusion}.

At the beginning of round $i$, we keep the feature extractor of the global model ($E_g$) instead of throwing it away as in FedAvg after initialization.
During training, $E_g$ is frozen and an additional feature fusion module described in Section \ref{sec:fusion} is introduced.
In practice, it's possible to record the global feature maps generated by $E_g$ in one round forward inference.
In other words, the additional feature fusion module brings limited extra computation costs.
After the on-device training procedure, the local model combined with the feature fusion module will be sent to the central server for model aggregation.
For \textit{multi} and \textit{single} operators, we use an exponential moving average strategy to smooth the update.

The pseudo code of FedFusion is shown in Algorithm \ref{algorithm}.

\section{Experiments}
\label{sec:exp}

In this section, we will first present the experimental setup (Section \ref{sec:setup}) and then show the results of FedMMD (Section \ref{sec:fedmmd}) and FedFusion (Section \ref{sec:fedfusion}) under several different settings.

\subsection{Experimental Setup}
\label{sec:setup}

\textbf{Datasets}

We use MNIST \cite{lecun1998gradient} and CIFAR10 \cite{krizhevsky2009learning} as basic datasets in our experiments.
We further proposed three types of data partitions to benchmark our FedMMD and FedFusion, and the vanilla FedAvg.

The first is \textit{Artificial non-IID Partition}, which is implemented by splitting an existing IID dataset to meet the FL settings and commonly used in previous FL studies \cite{konevcny2015federated,konevcny2016federated,mcmahan2017communication,caldas2018expanding,zhao2018federated}.
In this partition, a single client usually has a subset of the classes of the total data.
For example, most clients have up to two digits of MNIST in \cite{mcmahan2017communication}.

The second is \textit{User Specific non-IID Partition}, where the data on different clients usually have similar classes but follows different distributions.
This is commonly used in multi-task learning studies \cite{smith2017federated,chen2018federated,caldas2018leaf}.

The last is \textit{IID Partition}, which is a simple yet necessary partition to evaluate FL algorithms \cite{mcmahan2017communication,caldas2018expanding}.

\subsubsection{Models}

For MNIST digits recognition task, we use the the same model as FedAvg \cite{mcmahan2017communication}: a CNN with two 5$\times$5 convolution layers (the first with 32 channels while the second with 64, each followed by a ReLU activation and 2$\times$2 max pooling), a fully connected layer with 512 units (with a ReLU activation and random dropout), and a final softmax output layer.

For CIFAR10 we use a CNN with two 5$\times$5 convolution layers (both with 64 channels, each followed by a ReLU activation and 3$\times$3 max pooling with stride size 2), two fully connected layers (the first with 384 units while the second with 192, each followed by a ReLU activation and random dropout) and a final softmax output layer.

\subsection{FedMMD}
\label{sec:fedmmd}

\subsubsection{CNN on CIFAR-10}

\begin{figure}[t]
\centering
\begin{subfigure}[t]{0.49\linewidth}
    \centering
    \includegraphics[width=\linewidth]{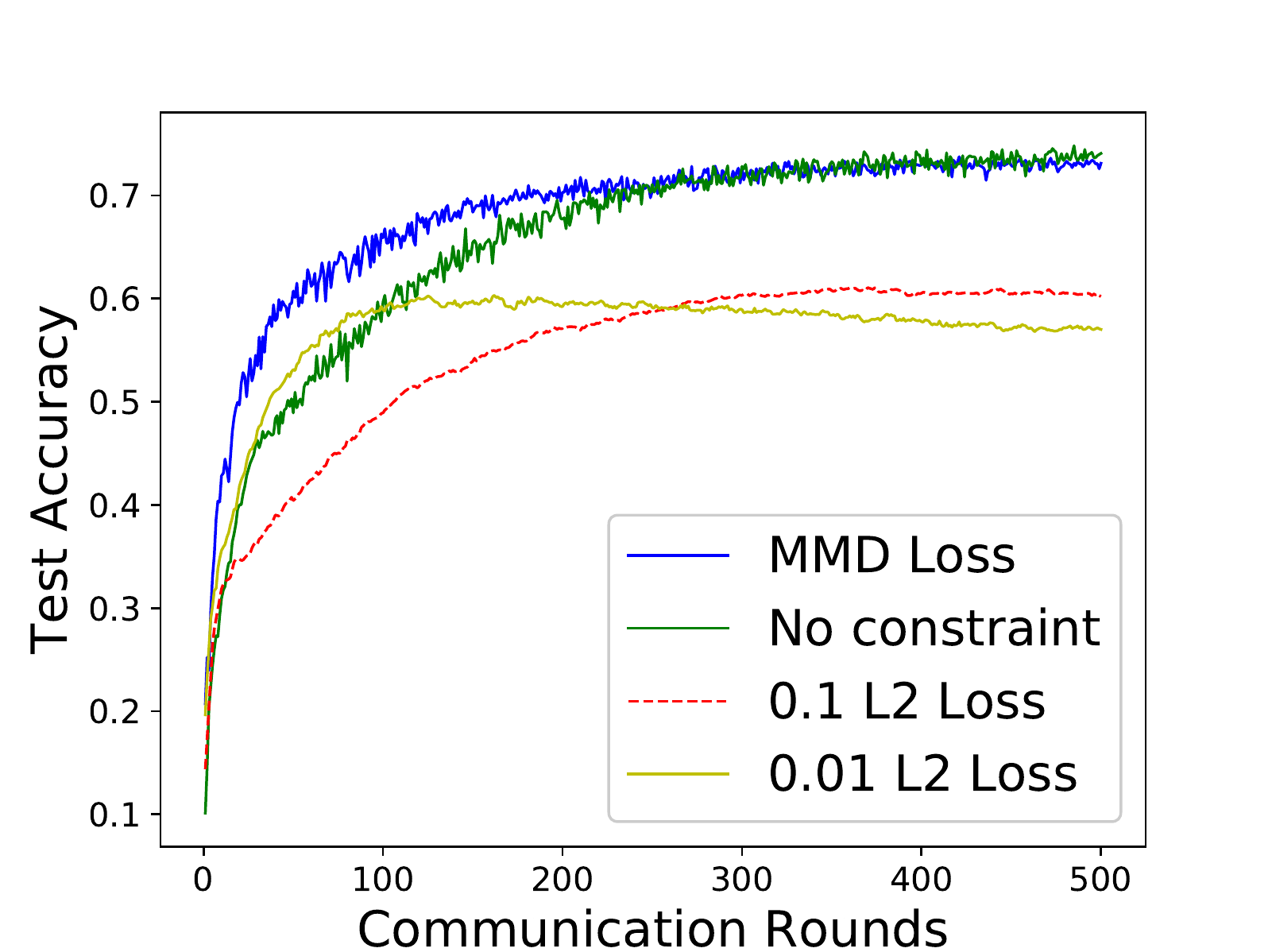}
    \caption{\label{fig:fedmmd_1} non-IID on CIFAR-10}
\end{subfigure}
\begin{subfigure}[t]{0.49\linewidth}
    \centering
    \includegraphics[width=\linewidth]{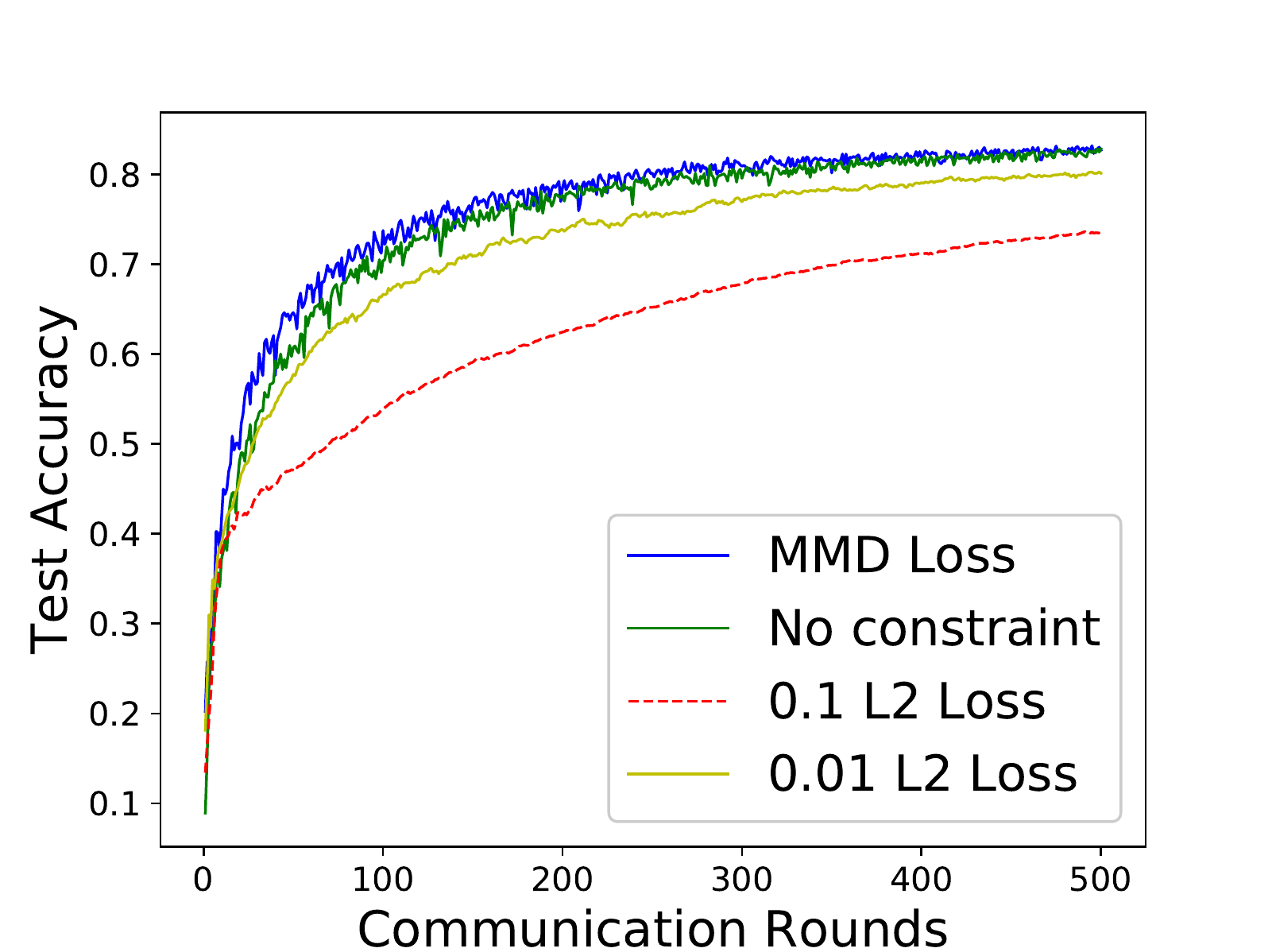}
    \caption{\label{fig:fedmmd_2} IID on CIFAR-10}
\end{subfigure}
\begin{subfigure}[t]{0.49\linewidth}
    \centering
    \includegraphics[width=\linewidth]{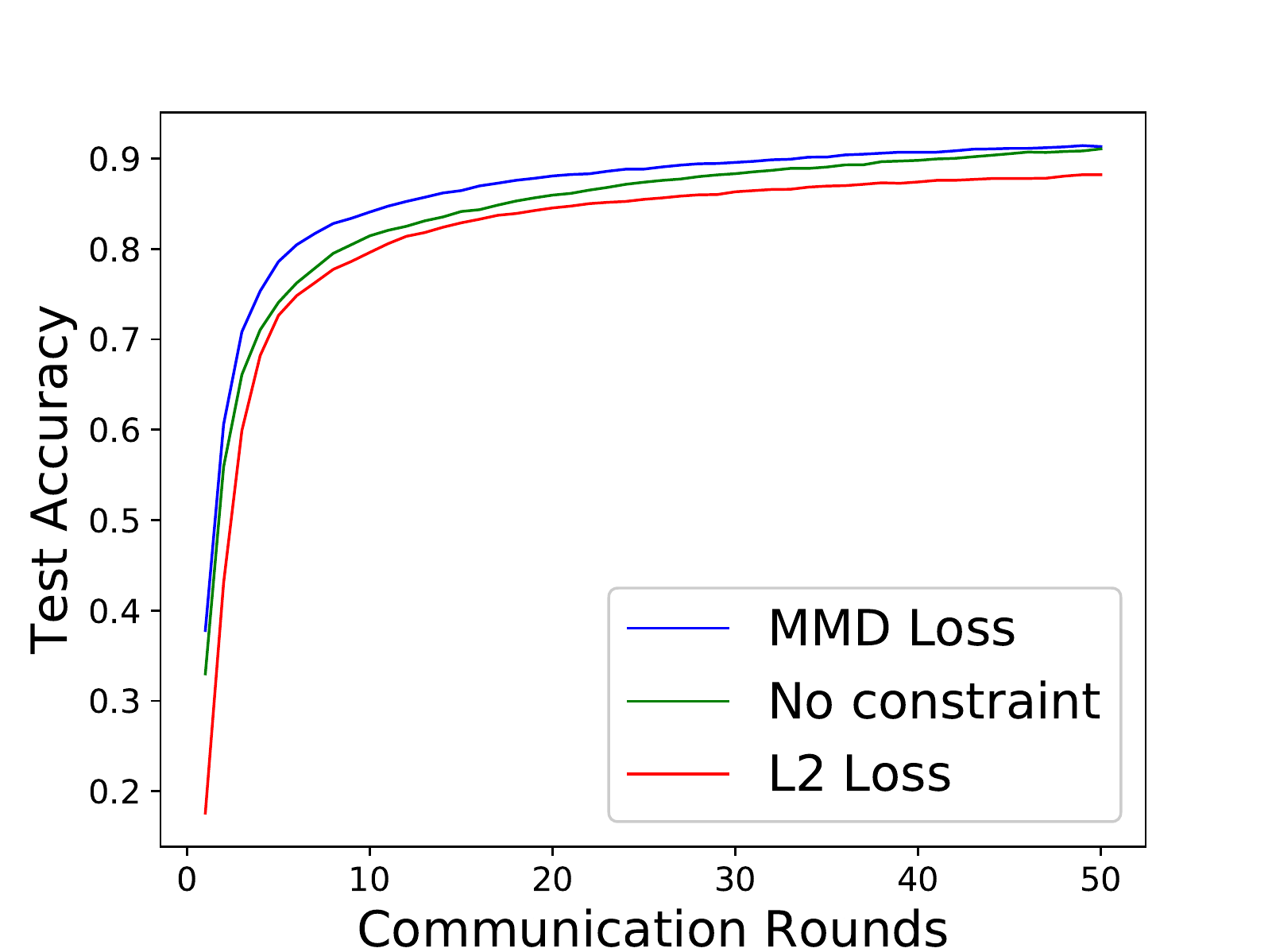}
    \caption{\label{fig:fedmmd_3} binary separated non-IID on MNIST}
\end{subfigure}
\begin{subfigure}[t]{0.49\linewidth}
    \centering
    \includegraphics[width=\linewidth]{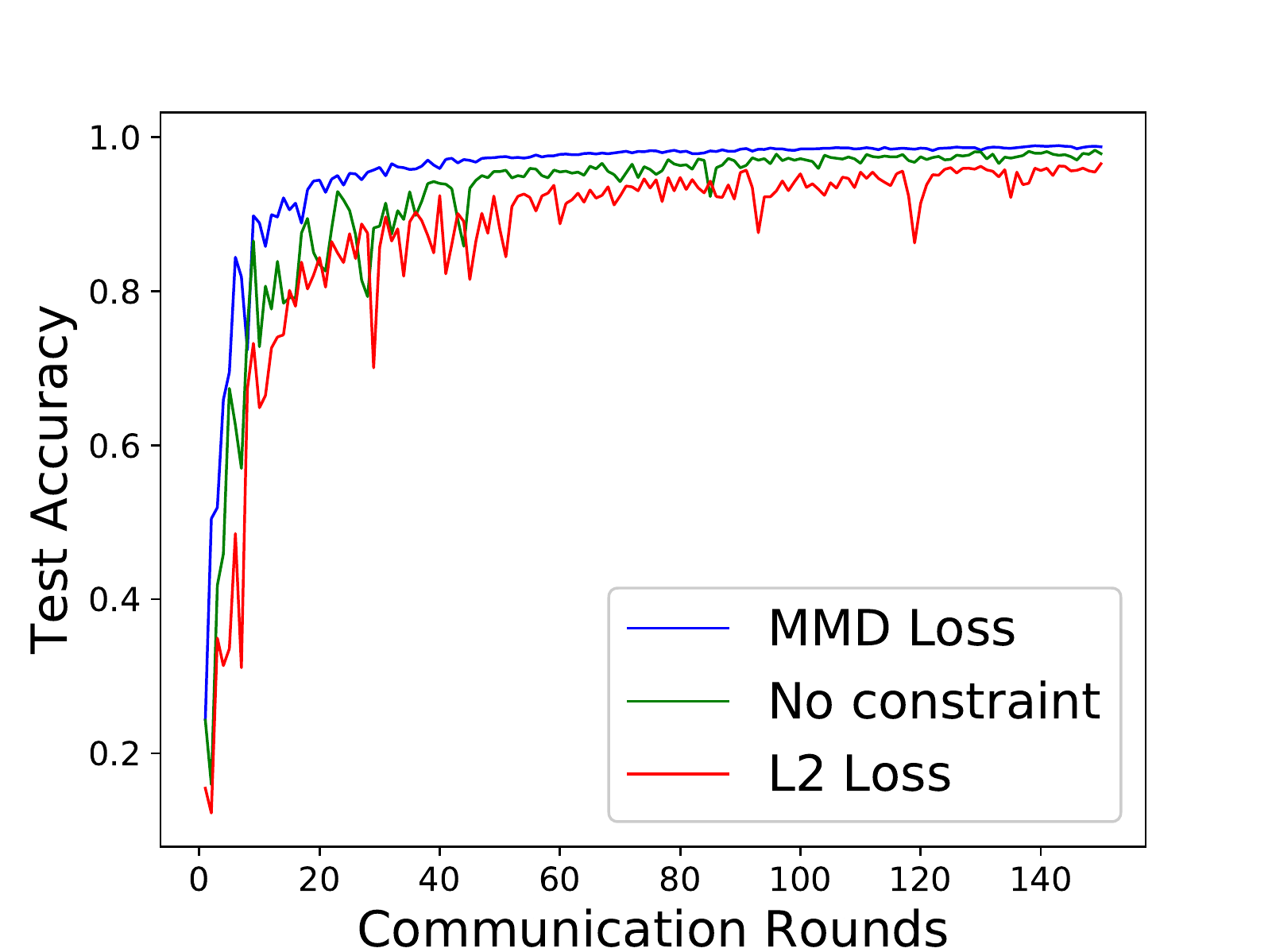}
    \caption{\label{fig:fedmmd_4} 100-clients non-IID on MNIST}
\end{subfigure}
\caption{Test accuracy over communication rounds for FedMMD on CIFAR-10 (upper row) and MNIST (lower row) with binary separated non-IID (left column) and 100-clients non-IID (right column) settings. (Better viewed in color)}
\label{fig:mmd}
\end{figure}

For convenience but without loss of generality, we select a fixed group of hyper-parameters, that is, two clients participating in the training process, with a local batch size $B=128$ and local epochs $E=2$.
During the training, we use a SGD optimizer with the learning  rate of $2 \times 10^{-3}$. 
The penalty parameter $\lambda$ for $L_{\mathit{MMD}}$ is set to 0.1 in this experiment.
As for the  penalty parameter for L2 norm, we have tried 0.1 and 0.01, and the results are shown in Figure \ref{fig:fedmmd_1}, \ref{fig:fedmmd_2}.

We explore both non-IID and IID data distributions and show that our FedMMD outperforms the baseline methods, especially in non-IID data distribution.

Figure \ref{fig:fedmmd_1} shows the non-IID situation, where we split the 10 classes of images in CIFAR-10 dataset into 2 parts, each containing 5 classes without overlap, indicating the non-IID data distribution.
FedMMD needs fewer communication rounds to get to convergence.
More concretely, FedMMD reaches the test accuracy of 0.72 by 260 rounds of communication, with a reduction of 20.2\% compared to 326 rounds need by FedAvg.
It is worth noting that, MMD forces the local model learn more knowledge from the global model but does not introduce new information into the overall optimization system compared with FedAvg, thus the final convergence results are the same, which is consistent as expected.

Figure \ref{fig:fedmmd_2} shows the IID situation.
As we can see, FedMMD performs similar to the vanilla FedAvg.
In this case, the data on each client is a uniform sample of the overall dataset, which means the local model is able to learn the complete representations, and as a result, the role of MMD constraint is weakened.
Two-stream model constrained by L2 norm underperforms other methods, indicating that selection of constraints does matter.

\subsubsection{CNN on MNIST}

Similar to the CIFAR-10 experiments, we first explore the binary separated non-IID distributions and the results are shown in Figure \ref{fig:fedmmd_3}.
During the training, we use a SGD optimizer with the learning rate of $1 \times 10^{-4}$.
The penalty parameter $\lambda$ for $L_{\mathit{MMD}}$ and $L_2$ are 0.1 and 0.001, respectively.
We use a rather small $\lambda$ for $L_2$ as we find that a larger one may lead to non-convergence situation. 
As we can see, our FedMMD are faster in convergence compared with other methods and does not lower the final test accuracy.
The reduction of communication rounds is less than the result on CIFAR-10.
We speculate that the variety of feature representations in MNIST is much less due to the black-and-white images and simple lines in them.

We further study a more complex non-IID data distribution described in \cite{mcmahan2017communication}, where we first sort the data in MNIST by digit label, divide it into 200 shards of size 300, and assign each of 100 clients 2 shards.
This is a typical non-IID partition of the data, as most clients will only have examples of two digits.
And we set $C=0.1$, with local batch size $B=10$ and local epochs $E=2$.
As shown in Figure \ref{fig:fedmmd_4}, the curve of test accuracy dithers due to the extremely pathological data distribution.
Our FedMMD achieves a test accuracy of 98\% with 72 rounds of communication while FedAvg needs 128 rounds, which means a reduction of 23.4\%.

\subsection{FedFusion}
\label{sec:fedfusion}

\begin{figure}[t]
\begin{subfigure}[b]{.49\linewidth}
    \centering
    \includegraphics[width=\linewidth]{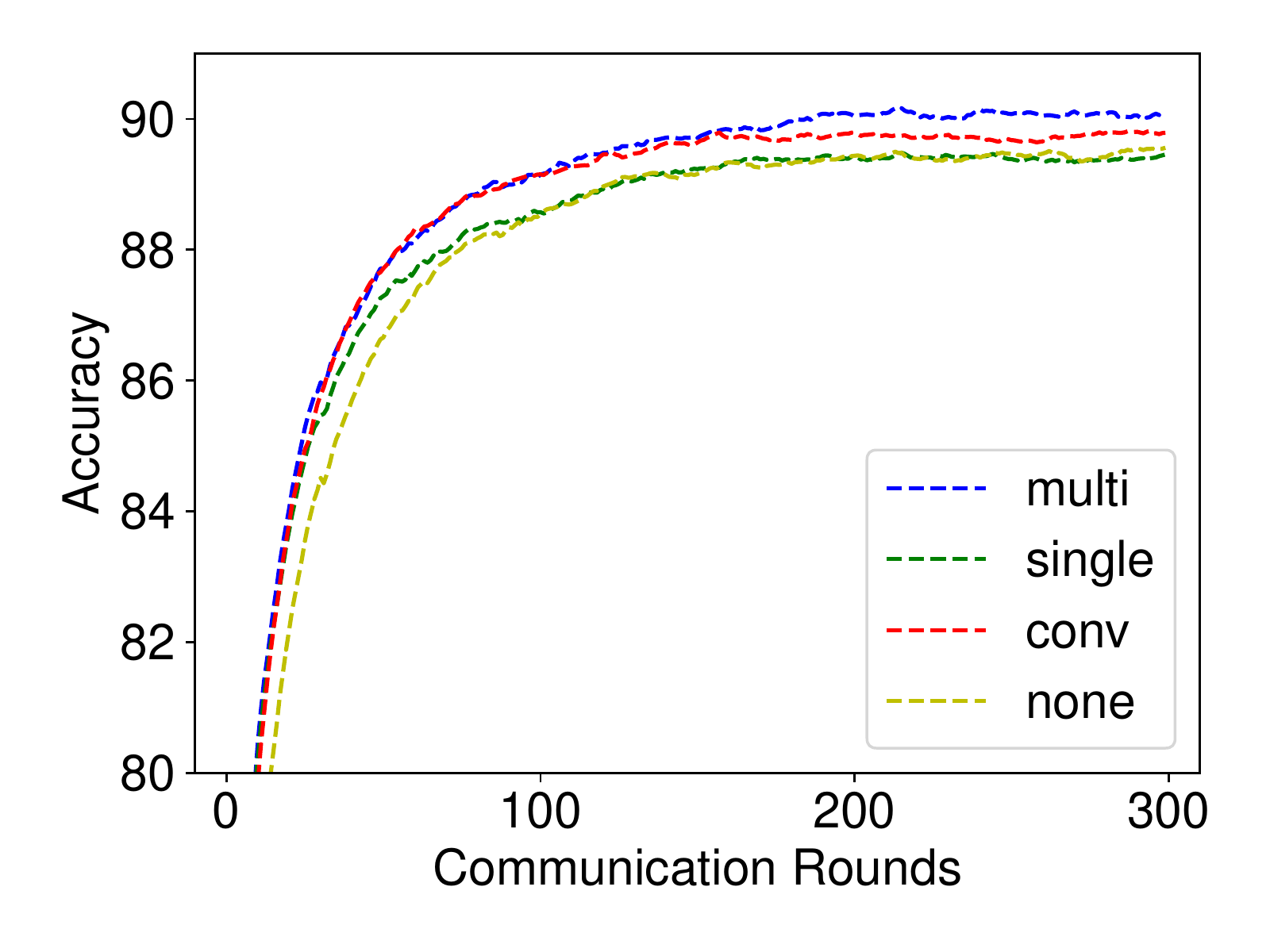}
    \caption{\label{fig:fusion_a} Artificial non-IID}
\end{subfigure}
\begin{subfigure}[b]{.49\linewidth}
    \centering
    \includegraphics[width=\linewidth]{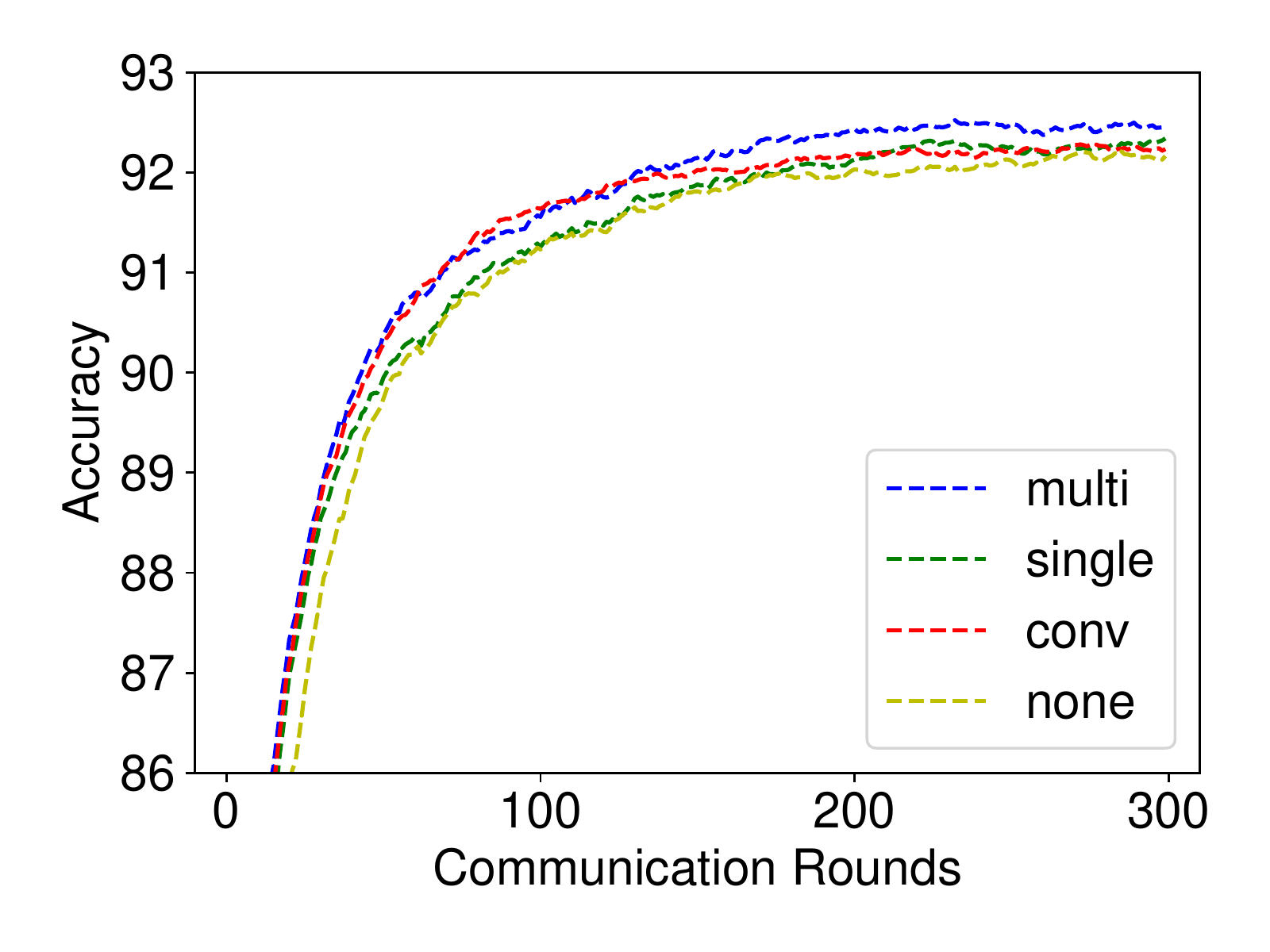}
    \caption{\label{fig:fusion_b} Artificial non-IID}
\end{subfigure}
\begin{subfigure}[b]{.49\linewidth}
    \centering
    \includegraphics[width=\linewidth]{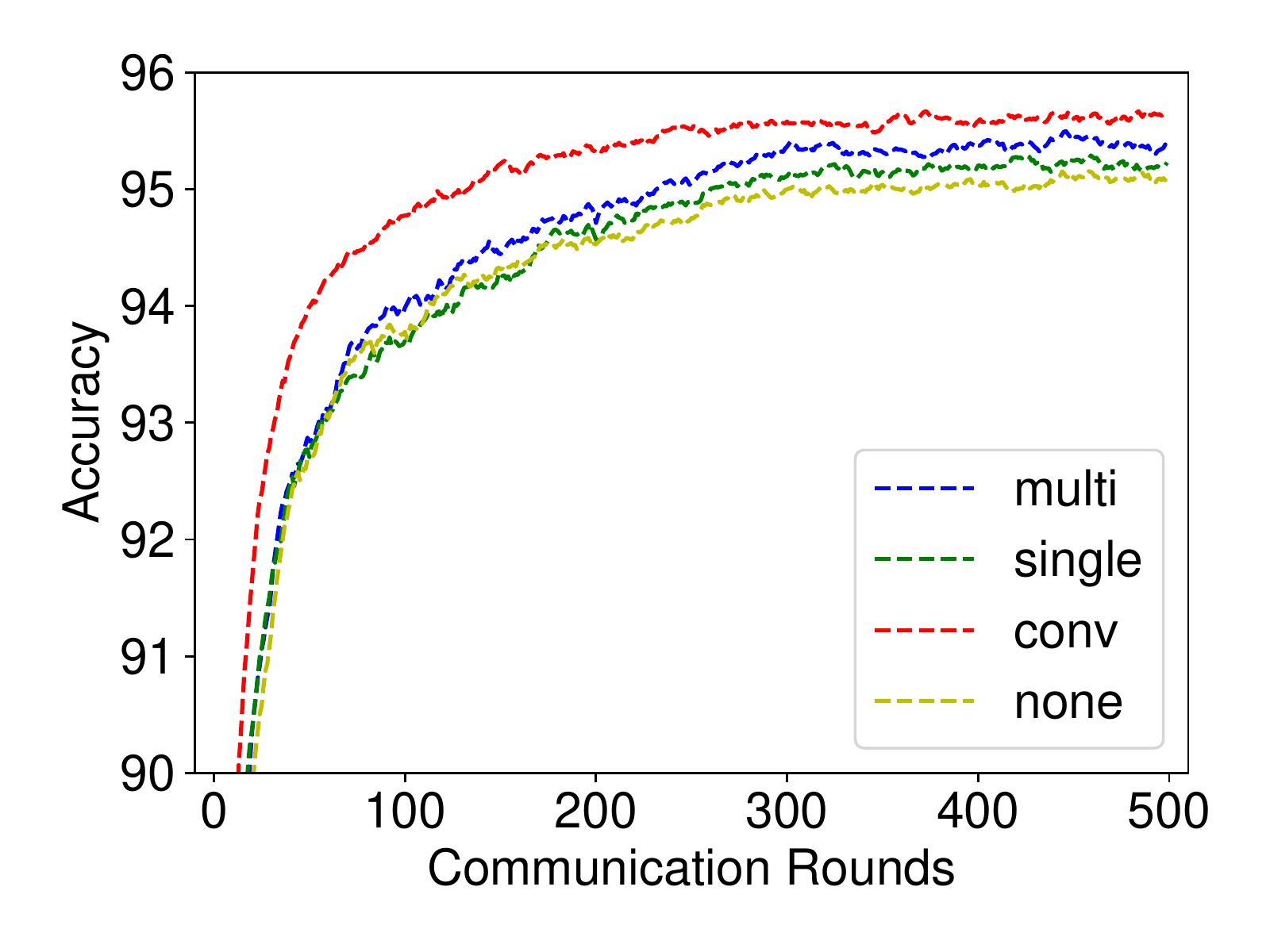}
    \caption{\label{fig:fusion_c} User-specific non-IID}
\end{subfigure}
\begin{subfigure}[b]{.49\linewidth}
    \centering
    \includegraphics[width=\linewidth]{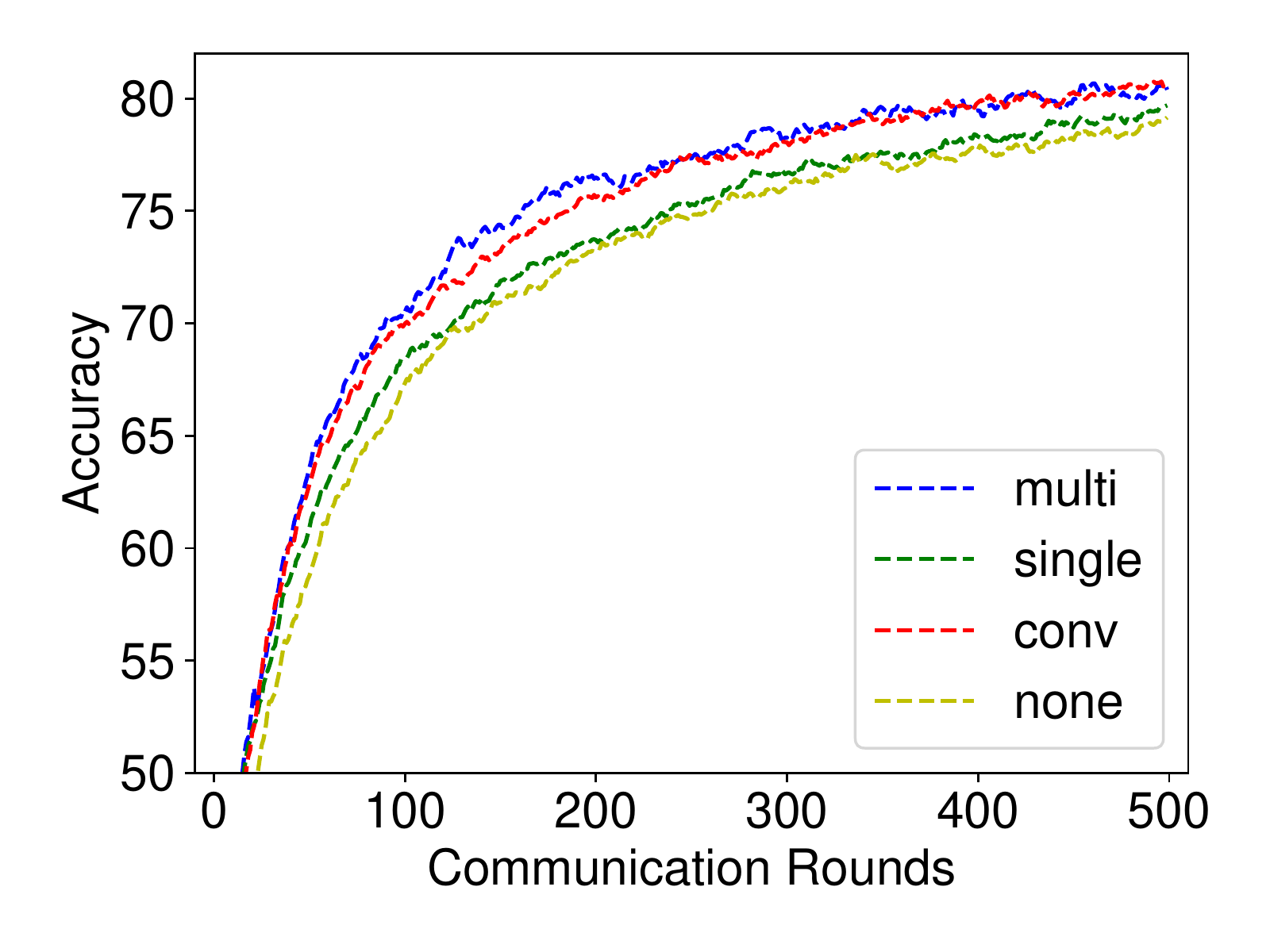}
    \caption{\label{fig:fusion_d} IID}
\end{subfigure}
\caption{Test accuracy vs. communication rounds under different settings.
(a) and (b): the artificial non-IID partitions of CIFAR10.
(c): user specific non-IID partition, which is implemented by applying different permutations to MNIST.
(d): IID partition of CIFAR10.
\textit{multi}, \textit{single} and \textit{conv} are FedFusion with corresponding fusion operators while \textit{none} denotes FedAvg.}
\label{fig:results}
\end{figure}

\subsubsection{Artificial non-IID Partition}

Under the artificial non-IID scenarios, we use a learning rate of 0.003 with an exponential decay factor 0.985 each round for all our FedFusion methods (with different fusion operators) and the compared FedAvg.
The convergence behaviors of them are illustrated in Figure \ref{fig:fusion_a} and \ref{fig:fusion_b} while the final accuracies at convergence are shown in Table \ref{tab:accuracy}.

The curve representing FedFusion with \textit{multi} operator is always above others, which means it achieves a higher accuracy at fewer communication costs.
The accuracy of FedFusion with \textit{conv} also raises faster at the beginning but fails to reach a better convergence point.
FedFusion with \textit{single} and FedAvg perform relatively worse.

Such results are obviously due to the \textit{multi} fusion operator.
As stated before, most clients have a subset of the total classes in artificial non-IID scenarios.
The \textit{multi} operator allows the models on clients to select the feature maps that are helpful to their local data.
In contrast, FedAvg does not offer the selection and the \textit{single} operator does not provide enough room for adjustment.

\subsubsection{User Specific non-IID Partition}

To simulate the \textit{user specific non-IID partition}, we apply different permutations to MNIST on each client, which is the so-called \textit{Permuted MNIST} in several previous studies \cite{goodfellow2013empirical, zenke2017continual}.
We use a learning rate of 0.002 with an exponential decay factor 0.99 each round for all the methods.

The number of communication rounds to reach certain accuracy milestones (94\% and 95\% here), as well as the reduction in communication rounds versus FedAvg, is shown in Table \ref{tab:rounds}.
The results indicate that FedFusion with \textit{conv} leads in a large margin, which is different from that in \textit{artificial non-IID partition}.
FedFusion with \textit{conv} achieves the best performance while reducing the number of communication rounds by more than 60\%.
In \textit{user specific non-IID partition}, the data on clients have similar classes but follow different distributions.
The \textit{conv} operator has better ability to integrate the feature maps from both the local and global models, in other words, the knowledge from other clients and data distributions.
It is worth noting that \textit{user specific non-IID partition} is closer to the realistic FL scenarios, thus the improvement makes more sense in this case.

\begin{table}
\caption{The convergence accuracy of FedFusion and FedAvg under different setups. (a-d) corresponds to those in Figure \ref{fig:results}.}
\label{tab:accuracy}
\centering
\begin{tabular}{@{}lllll@{}}
\toprule
                    & (a)            & (b)            & (c)            & (d)            \\ \midrule
FedAvg           & 89.89          & 92.21          & 95.20          & 80.01          \\ 
FedFusion+Single & 89.77          & 92.32          & 95.25          & 80.85          \\ 
FedFusion+Multi  & \textbf{90.51} & \textbf{92.78} & 95.43          & \textbf{82.95} \\ 
FedFusion+Conv   & 90.11          & 92.53          & \textbf{95.79} & 82.15          \\ \bottomrule
\end{tabular}
\end{table}

\begin{figure}[t]
\centering
\includegraphics[width=0.65\linewidth]{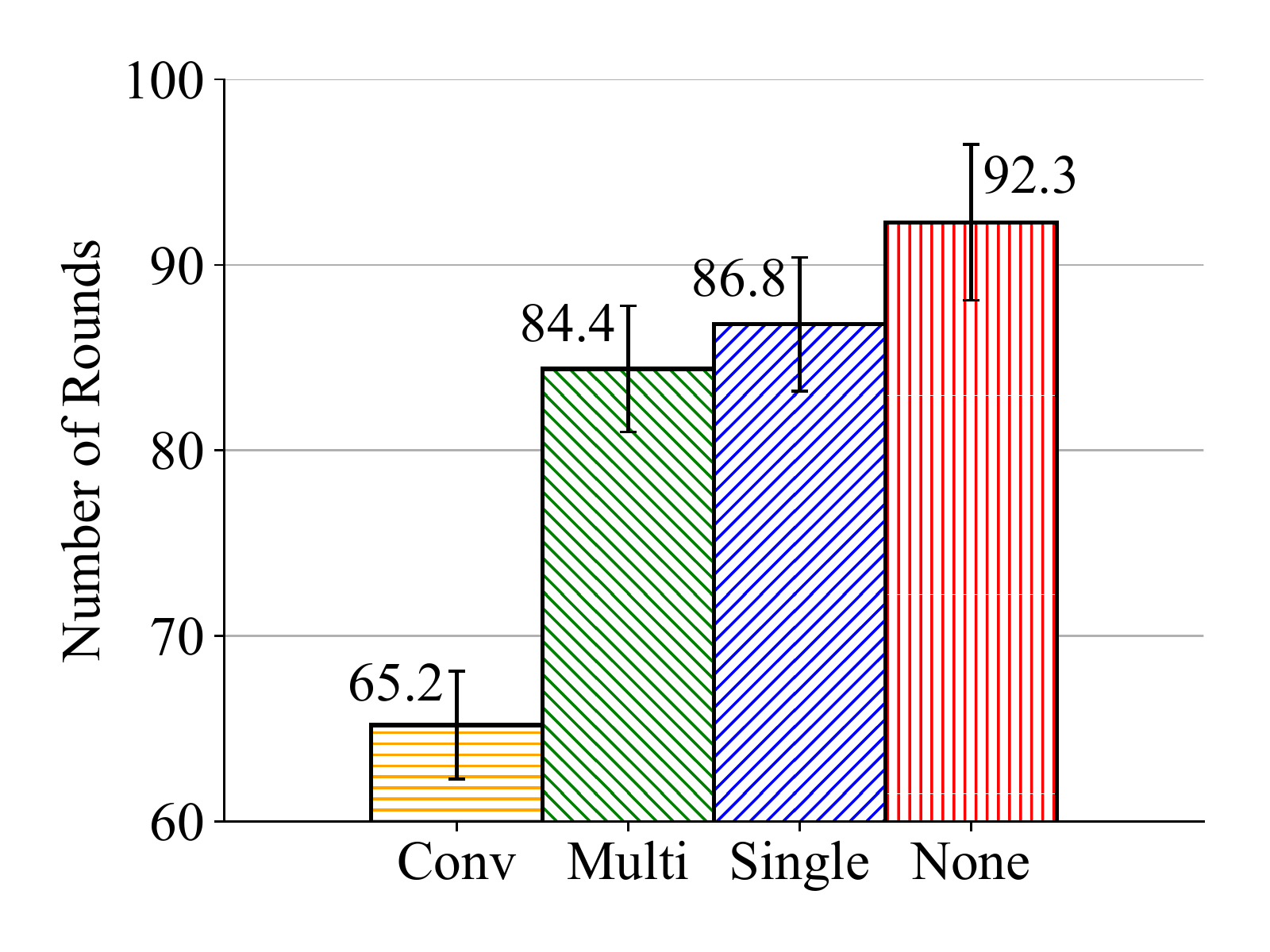}
\caption{Number of local epochs to reach convergence for newly incoming clients.}
\label{fig:rounds}
\end{figure}

Additionally, we study the generalization ability of the model that was usually ignored in previous FL research works.
The local epochs to reach convergence for newly incoming clients are illustrated in Figure \ref{fig:rounds}.
As we can see, when a new client joins an existing FL system, FedFusion with \textit{conv} provides a better initialization than other algorithms and thus speeds up the process of convergence.

\begin{table}
\caption{Number of communication rounds to reach certain accuracy milestones. FedAvg is considered as the baseline, and reductions in communication rounds are listed.}
\label{tab:rounds}
\centering
\begin{tabular}{@{}llllll@{}}
\toprule
\multirow{2}{*}{} & \multicolumn{2}{c}{94\%}     & & \multicolumn{2}{c}{95\%}      \\ \cmidrule{2-3} \cmidrule{5-6} 
                    & rounds      & reduce         & & rounds      & reduce          \\ \midrule
FedAvg            & 100         & (ref)          & & 256         & (ref)           \\
FedFusion+Single  & 87          & 13.0\%         & & 227         & 11.3\%          \\
FedFusion+Multi   & 78          & 22.0\%         & & 201         & 21.5\%          \\
FedFusion+Conv    & \textbf{34} & \textbf{66.0\%}& & \textbf{92} & \textbf{64.1\%} \\ \bottomrule
\end{tabular}
\end{table}

\subsubsection{IID Partition}

The \textit{IID partition} is a simple yet necessary partition to evaluate FL algorithms.
If one strategy cannot handle this partition, its effectiveness is questionable.

As shown in Figure \ref{fig:fusion_d}, FedFusion with \textit{multi} and \textit{conv} achieve higher accuracy at fewer communication costs.
In terms of the final convergence accuracy, FedFusion with \textit{multi} and \textit{conv} have an impressive improvement than other methods.

To make a brief conclusion on the feature fusion operators as follows:
The \textit{multi} operator offers flexible choices between the local and global feature maps and is more interpretable.
Entries in the weight vector $\pmb{\lambda}$ account for the proportions of the global feature maps in corresponding channels.
When there were gaps in the classes of data, \textit{multi} operators would learn to choose the most helpful feature maps.
The \textit{conv} operator is better at integrating the knowledge from the global and local models.
If the data on clients had similar classes but followed different distributions, \textit{conv} operator performs much better.
Our experiments indicate that \textit{single} operator has few improvements and should not be adopted in practice.

\section{Conclusion}
\label{sec:conclusion}

The heavy communication costs of FedAvg is an emergency problem to solve.
In this paper, we first replace the single model trained on clients in FL settings with a two-stream model consisting of the global and local ones.
Our experiments show that introducing the MMD constraint into the optimization algorithm will bring a reduction in communication rounds, especially in non-IID FL settings.
Further we propose a new FL algorithm with feature fusion modules and evaluate it in popular FL setups.
The experimental results show that the proposed method achieves a higher accuracy while reducing the communication rounds by more than 60\%.
What is more, we observe that FedFusion offers better generalization for newly incoming clients.

\section{Acknowledgement}
This work is supported by the National Key R\&D Program of China (2018YFB1003703), National Natural Science Foundation of China (61472204 \& 61521002), as well as Beijing Key Lab of Networked Multimedia (Z161100005016051).

\clearpage
\bibliographystyle{abbrv}
\bibliography{ref.bib}

\begin{thebibliography}{10}

\bibitem{caldas2018expanding}
S.~Caldas, J.~Kone{\v{c}}ny, H.~B. McMahan, and A.~Talwalkar.
\newblock Expanding the reach of federated learning by reducing client resource
  requirements.
\newblock {\em arXiv preprint arXiv:1812.07210}, 2018.

\bibitem{caldas2018leaf}
S.~Caldas, P.~Wu, T.~Li, J.~Kone{\v{c}}n{\`y}, H.~B. McMahan, V.~Smith, and
  A.~Talwalkar.
\newblock Leaf: A benchmark for federated settings.
\newblock {\em arXiv preprint arXiv:1812.01097}, 2018.

\bibitem{chen2018federated}
F.~Chen, Z.~Dong, Z.~Li, and X.~He.
\newblock Federated meta-learning for recommendation.
\newblock {\em arXiv preprint arXiv:1802.07876}, 2018.

\bibitem{goodfellow2013empirical}
I.~J. Goodfellow, M.~Mirza, D.~Xiao, A.~Courville, and Y.~Bengio.
\newblock An empirical investigation of catastrophic forgetting in
  gradient-based neural networks.
\newblock {\em arXiv preprint arXiv:1312.6211}, 2013.

\bibitem{gretton2012optimal}
A.~Gretton, D.~Sejdinovic, H.~Strathmann, S.~Balakrishnan, M.~Pontil,
  K.~Fukumizu, and B.~K. Sriperumbudur.
\newblock Optimal kernel choice for large-scale two-sample tests.
\newblock In {\em Advances in neural information processing systems}, pages
  1205--1213, 2012.

\bibitem{jeong2018communication}
E.~Jeong, S.~Oh, H.~Kim, J.~Park, M.~Bennis, and S.-L. Kim.
\newblock Communication-efficient on-device machine learning: Federated
  distillation and augmentation under non-iid private data.
\newblock {\em arXiv preprint arXiv:1811.11479}, 2018.

\bibitem{konevcny2015federated}
J.~Kone{\v{c}}n{\`y}, B.~McMahan, and D.~Ramage.
\newblock Federated optimization: Distributed optimization beyond the
  datacenter.
\newblock {\em arXiv preprint arXiv:1511.03575}, 2015.

\bibitem{konevcny2016federated}
J.~Kone{\v{c}}n{\`y}, H.~B. McMahan, F.~X. Yu, P.~Richt{\'a}rik, A.~T. Suresh,
  and D.~Bacon.
\newblock Federated learning: Strategies for improving communication
  efficiency.
\newblock {\em arXiv preprint arXiv:1610.05492}, 2016.

\bibitem{krizhevsky2009learning}
A.~Krizhevsky and G.~Hinton.
\newblock Learning multiple layers of features from tiny images.
\newblock Technical report, Citeseer, 2009.

\bibitem{lecun1998gradient}
Y.~LeCun, L.~Bottou, Y.~Bengio, and P.~Haffner.
\newblock Gradient-based learning applied to document recognition.
\newblock {\em Proceedings of the IEEE}, 86(11):2278--2324, 1998.

\bibitem{long2015learning}
M.~Long, Y.~Cao, J.~Wang, and M.~I. Jordan.
\newblock Learning transferable features with deep adaptation networks.
\newblock {\em arXiv preprint arXiv:1502.02791}, 2015.

\bibitem{long2013transfer}
M.~Long, J.~Wang, G.~Ding, J.~Sun, and S.~Y. Philip.
\newblock Transfer feature learning with joint distribution adaptation.
\newblock In {\em Computer Vision (ICCV), 2013 IEEE International Conference
  on}, pages 2200--2207. IEEE, 2013.

\bibitem{mcmahan2017communication}
B.~McMahan, E.~Moore, D.~Ramage, S.~Hampson, and B.~A. y~Arcas.
\newblock Communication-efficient learning of deep networks from decentralized
  data.
\newblock In {\em Artificial Intelligence and Statistics}, pages 1273--1282,
  2017.

\bibitem{smith2017federated}
V.~Smith, C.-K. Chiang, M.~Sanjabi, and A.~S. Talwalkar.
\newblock Federated multi-task learning.
\newblock In {\em Advances in Neural Information Processing Systems}, pages
  4424--4434, 2017.

\bibitem{sun2016deep}
B.~Sun and K.~Saenko.
\newblock Deep coral: Correlation alignment for deep domain adaptation.
\newblock In {\em European Conference on Computer Vision}, pages 443--450.
  Springer, 2016.

\bibitem{suresh2017distributed}
A.~T. Suresh, F.~X. Yu, S.~Kumar, and H.~B. McMahan.
\newblock Distributed mean estimation with limited communication.
\newblock In {\em Proceedings of the 34th International Conference on Machine
  Learning-Volume 70}, pages 3329--3337. JMLR. org, 2017.

\bibitem{tzeng2014deep}
E.~Tzeng, J.~Hoffman, N.~Zhang, K.~Saenko, and T.~Darrell.
\newblock Deep domain confusion: Maximizing for domain invariance.
\newblock {\em arXiv preprint arXiv:1412.3474}, 2014.

\bibitem{yao2018twostream}
X.~Yao, C.~Huang, and L.~Sun.
\newblock Two-stream federated learning: Reduce the communication costs.
\newblock In {\em Visual Communications and Image Processing (VCIP), 2018},
  pages 1--4. IEEE, 2018.

\bibitem{yao2019towards}
X.~Yao, T.~Huang, C.~Wu, R.~Zhang, and L.~Sun.
\newblock Towards faster and better federated learning: A feature fusion
  approach.
\newblock In {\em IEEE International Conference on Image Processing}, 2019.

\bibitem{zenke2017continual}
F.~Zenke, B.~Poole, and S.~Ganguli.
\newblock Continual learning through synaptic intelligence.
\newblock In {\em International Conference on Machine Learning}, pages
  3987--3995, 2017.

\bibitem{zhao2018federated}
Y.~Zhao, M.~Li, L.~Lai, N.~Suda, D.~Civin, and V.~Chandra.
\newblock Federated learning with non-iid data.
\newblock {\em arXiv preprint arXiv:1806.00582}, 2018.

\bibitem{zhuo2017deep}
J.~Zhuo, S.~Wang, W.~Zhang, and Q.~Huang.
\newblock Deep unsupervised convolutional domain adaptation.
\newblock In {\em Proceedings of the 2017 ACM on Multimedia Conference}, pages
  261--269. ACM, 2017.

\end{thebibliography}

\end{document}